\documentclass[nohyperref]{article}

\usepackage{microtype}
\usepackage{graphicx}
\usepackage[T1]{fontenc}
\usepackage{tabularx}
\usepackage{amsmath}
\usepackage{caption}
\usepackage{booktabs} 

\usepackage{hyperref}

\usepackage[accepted]{icml2022}
\usepackage{subfigure}

\usepackage{amsmath}
\usepackage{amssymb}
\usepackage{mathtools}
\usepackage{amsthm}

\usepackage[capitalize,noabbrev]{cleveref}

\theoremstyle{plain}

\theoremstyle{definition}

\theoremstyle{remark}

\usepackage[textsize=tiny]{todonotes}

\icmltitlerunning{Submission and Formatting Instructions for ICML 2022}

\begin{document}

\twocolumn[
\icmltitle{A Study on the Predictability of Sample Learning Consistency}



\icmlsetsymbol{equal}{*}

\begin{icmlauthorlist}
\icmlauthor{Alain Raymond-Sáez}{puc,cenia}
\icmlauthor{Julio Hurtado}{pisa}
\icmlauthor{Álvaro Soto}{puc}
\end{icmlauthorlist}

\icmlaffiliation{puc}{Department of Computer Science, Pontificia Universidad Católica de Chile, Santiago, Chile}
\icmlaffiliation{pisa}{Department of Computer Science, University of Pisa, Pisa, Italy}
\icmlaffiliation{cenia}{Centro Nacional de Inteligencia Artificial (CENIA), Chile}
\icmlcorrespondingauthor{Alain Raymond}{afraymon@uc.cl}

\icmlkeywords{Machine Learning, ICML}

\vskip 0.3in
]



\printAffiliationsAndNotice{\icmlEqualContribution} 

\begin{abstract}
Curriculum Learning is a powerful training method that allows for faster and better training in some settings. This method, however, requires having a notion of which examples are difficult and which are easy, which is not always trivial to provide. A recent metric called C-Score acts as a proxy for example difficulty by relating it to learning consistency. Unfortunately, this method is quite compute intensive which limits its applicability for alternative datasets. In this work, we train models through different methods to predict C-Score for CIFAR-100 and CIFAR-10. We find, however, that these models generalize poorly both within the same distribution as well as out of distribution. This suggests that C-Score is not defined by the individual characteristics of each sample but rather by other factors. We hypothesize that a sample's relation to its neighbours, in particular, how many of them share the same labels, can help in explaining C-Scores. We plan to explore this in future work.


\end{abstract}
\vspace{-0.5cm}
\section{Introduction}
\vspace{-0.1cm}
\label{Introduction}

While Deep Learning has achieved significant milestones over the last decade, training strategies have remained relatively unchanged.
Over the last years, strides have been made in improving these strategies. One such approach is Curriculum Learning (CL) \citep{10.1145/1553374.1553380, wu2021when}. CL strategies are based on providing a model with an increasingly difficult selection of examples, similar to how children learn in school. CL strategies are popular in Reinforcement Learning -where a curriculum of tasks is progressively learned - and Natural Language Processing applications - where commonly sentence length has been a proxy for difficulty.

The problem, however, lies in how to determine the difficulty of samples. Typically, an \textit{ad hoc} difficulty measure needs to be crafted by hand for a specific problem, which limits their application in practice. Even worse, for some applications like image classification, sometimes there's no natural difficulty ordering that can be created. This has led to the development of more general measures of difficulty: Self Paced Learning\cite{Kumar2010} utilizes the current loss of an example as a proxy for difficulty; however, this requires extra computation and is prone to overfitting\cite{Jiang2015}. As an alternative, a metric called C-Score was proposed \citep{DBLP:conf/icml/JiangZTM21}. This metric measures how consistently a particular example is learned across various models, that is, it measures the proportion of iterations where a sample is successfully classified. The logic behind this metric is supported by the empirical observation that samples are generally learned in the same order \citep{DBLP:conf/iclr/TonevaSCTBG19} -independent of the model's architecture- and that a significant number of examples are learned once and never forgotten. The drawback to this metric is that it requires inordinate amounts of computing for a single dataset, as it requires several models to be trained and averaged over, negating many of the practical benefits of applying CL in the first place.


A reasonable strategy to alleviate the problems of computing C-Scores would be to learn a model that could extrapolate C-Scores to unseen datasets. This could be done by training a model on datasets where the score is already calculated.  In order for this strategy to be successful, C-Scores should be mainly determined by each sample's features. However, we know C-Scores must necessarily depend also on the labels associated to the dataset, as it has been seen\cite{pmlr-v70-arpit17a} that natural images with random labels are harder to learn. This leads us to ask the following research questions:

\begin{itemize}
    \item How much does C-Score depend on a sample's features? How much does it depend on the task?  
    \item Is there enough information in a sample's features to predict reasonable estimates of C-Score for similar unseen tasks?
    \item If so, are these estimates capable of creating useful curricula for unseen tasks?
\end{itemize}

To answer these questions, we attempt to learn models that predict C-Score through a variety of different methods. We find, however, that all methods produce models that generalize poorly even in the same dataset, while they barely generalize above random chance to unseen datasets.

Our current findings suggest that C-Score cannot be explained solely by a sample's features, even when training multiple models on this task. Our current hypothesis, which we plan to explore in future work, is that understanding how a sample relates to its neighbours, in particular, whether those neighbours belong to the same class or not can shed some light into explaining C-Scores, as well.


\section{Methodology}

\begin{figure}[h]
    \centering
    \subfigure[CIFAR-10]{\includegraphics[width=0.4\textwidth]{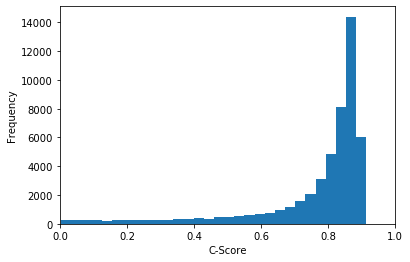}}
    \subfigure[CIFAR-100]{\includegraphics[width=0.4\textwidth]{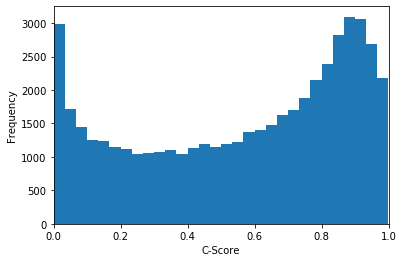}}
\caption{C-Score Distribution for datasets used in our experiments.}
\label{fig:hist_c_score}
\vspace{-2mm}
\end{figure}

C-Score provides a measure of how consistently a sample is learned by a model. In particular, it defines the average ratio of iterations for which a group of models correctly predict the label of a particular sample. Therefore, it is defined in the range [0,1], with higher values indicating easier to learn examples.

We attempt to predict C-Scores by three different methods: regression, Bayesian Personalized Ranking (BPR)\cite{rendle2012bayesian} and  classification by binning C-Scores. These methods were selected as they provide different relaxations to the original problem with regression being the hardest and binning the simplest version of the same problem.

\subsection{Experiment Setup}
We use ResNet18\cite{DBLP:conf/cvpr/HeZRS16} as our architecture of choice. Each experiment is run for 200 epochs using a learning rate of 0.001, with 1 epoch warmup. We use learning rate scheduling with a decay rate of $0.2$ at 60, 120, 160 epochs. We use weight decay of $5e4$. All experiments use batch size 256. We performed experiments with both pretrained models and randomly initialized models. All results correspond to randomly initialized models as they performed equal or better than pretrained models. All experiments are run using 10 different seeds.

For Regression and BPR, we change the last classification layer for a linear layer with one output. Adding a sigmoid activation function after this layer made no difference in performance, so we skip them in reported experiments.

\subsection{Datasets}
We work with CIFAR-10\cite{CIFAR10} and CIFAR-100\cite{CIFAR100} as these are the datasets for which there are published C-Scores. As C-Scores are only defined for the training sets of these datasets, we partition them into training and test sets by a random 80/20 split. As shown in Figure \ref{fig:hist_c_score}, CIFAR-10 shows a skewed distribution of C-Scores, with most scores being around 0.8, while CIFAR-100 and ImageNet show a more uniform distribution of scores, with peaks near 0 and 1.

\subsection{Regression}
We will start by trying to learn models that directly regress C-Score based on an input image. We use Mean Squared Error (MSE) as our loss function.

\subsection{Bayesian Personalized Ranking}
BPR was originally conceived as a method to learn how to rank items in Recommender Systems, where it is useful to generate sequences of recommendations and learning how to do it from actual user preferences. This translates perfectly to our need to learn how to produce difficulty orderings that relate to the ground truth C-Scores that we know.
To achieve this, BPR defines a loss function that encourages the model to learn how to rank a pair of samples appropriately. The intuition behind this method is that instead of directly learning how to regress a score, we learn how to score such that for any given pair of examples the ordering is correct. 
In particular, given a pair of samples $x_1, x_2$,  and an ordering function $O(x)$ where $O(x_2) > O(x_1)$ the BPR loss is traditionally defined as:
\begin{equation*}
BPR(x_1, x_2) = -log(\sigma(x_2-x_1))
\end{equation*}During our experiments, we found this function to diverge and subsequently removed the logarithm. Thus our modified BPR loss is:
\begin{equation*}
BPR_m(x_1, x_2) = \sigma(x_1-x_2)
\end{equation*}In order to train with this loss, we simply forward propagate the sampled batch, we produce every possible pair between samples from that batch, calculate the BPR loss for each pair and backpropagate. With a batch size of 256, this amounts to 32.640 pairs per batch.
\vspace{-0.1cm}
\subsection{Binning}
\vspace{-0.1cm}
As direct regression might prove too difficult a task to learn, we attempt to train a model on a relaxed version of the problem. First, we divide the C-Score range of [0,1] in equal width bins. Then, we train a model to learn to classify which bin a certain image belongs to given a traditional Cross Entropy Loss. We run experiments with 5, 10, 20, 40 bins. Given the distribution of C-Scores, we tried training models weighting losses to take into account these imbalances. Results were no better than regular training, thus we will not report on those results.
\vspace{-0.1cm}
\subsection{Evaluation}
\vspace{-0.1cm}
All models are trained on CIFAR-100 and evaluated on two test sets: the CIFAR-100 test set and CIFAR-10. CIFAR-100 is meant to test in distribution generalization, while CIFAR-10 is meant to test out of distribution generalization.

In the case of BPR, as it is not computationally feasible to generate every possible test pair, we do an approximation by doing a fixed permutation of the test set and evaluating all pairs of the retrieved batches (of size 512).

Finally, we evaluate all models via Spearman Rank Correlation (SRC) to understand the quality of the orderings produced by each model. When using Binning, the ground truth ordering is defined by the bin to which each sample belongs. 

\begin{figure}[]{\includegraphics[width=0.45\textwidth]{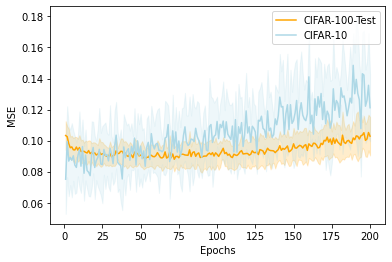}}
\caption{Test MSE for a Regression Model trained on CIFAR-100 C-Scores. In distribution performance remains stable, while out of distribution generalization degrades over time.}
\label{fig:regression}
\vspace{-4mm}
\end{figure}

\begin{figure}[h]
\subfigure[Test Performance on CIFAR-100]{\includegraphics[width=0.45\textwidth]{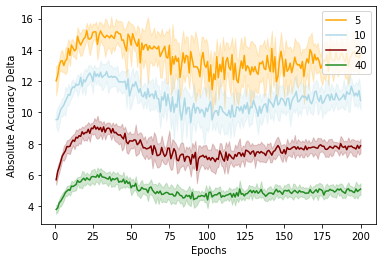}}
\subfigure[Out of Distribution Generalization on CIFAR-10]{\includegraphics[width=0.45\textwidth]{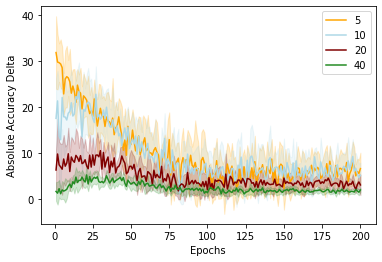}} 
\caption{Generalization performance increase over performance of a random baseline using Bins of size \{5,10,20,40\} training with CIFAR-100 C-Scores. As can be seen, performance within the same task achieves poor generalization. Out of distribution generalization barely surpasses random guessing.}
\label{fig:binning}
\vspace{-6mm}
\end{figure}

\begin{figure}[]{\includegraphics[width=0.45\textwidth]{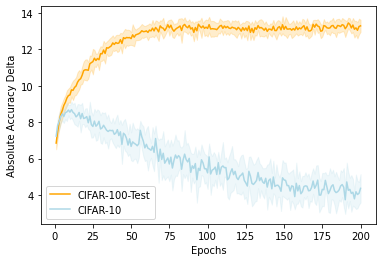}}
\caption{Generalization performance increase over performance of a random baseline using BPR training with CIFAR-100 C-Scores. A Random Baseline should achieve 50\% accuracy. As can be seen, performance within the same task achieves poor generalization. Out of distribution generalization barely surpasses random guessing. This suggests that image features on their own, do little to explain sample learning consistency.}
\label{fig:bpr}
\vspace{-5mm}
\end{figure}
\vspace{-0.2cm}
\section{Results}
\vspace{-0.2cm}
As can be seen in Figure \ref{fig:regression}, direct regression of the C-Score seems to work poorly with MSEs around 0.1. This implies an average error of around 0.3 when generalizing to the test set of CIFAR-100. Matters become worse when evaluating generalization to CIFAR-10, with MSE values reaching 0.16 which imply an average error of 0.3. This is significant when predicting scores in the C-Score range [0,1], suggesting poor prediction performance. Binning results are shown in Figure \ref{fig:binning}. The same pattern holds: generalization in CIFAR-100 is far greater than in CIFAR-10. CIFAR-10 performance is barely above random chance for all number of bins. This might be due to CIFAR-10's C-Scores skew towards values around 0.8. Increasing the number of bins rapidly degrades performance (as expected of a harder task). This decrease, however, is quite extreme showing that models learn hypotheses that do not generalize well. For BPR (Figure \ref{fig:bpr}), we observe the same behaviour, with CIFAR-10 and CIFAR-100 performance degrading to 4\% and 14\% above random chance.

For all methods, generalization capabilities \textit{within the same dataset} are quite low, while extrapolation to a new task is barely above random chance, which suggests that the task may require additional information to be solved. One could argue, however, that an exact prediction is not exactly what is needed, just a reasonable ordering. Thus while we may have poor models at predicting exact scores, they may in the end produce decent orderings. To test this, we calculate the Spearman Rank Correlation (SRC) between predictions of all our models against the ground truth. Results are shown in Table \ref{sample-table}. Correlations within the same dataset move between 0.28 and 0.44, with BPR being the best performing method of them all. This is a reasonable amount of correlation, but hardly explains the C-Score ordering. This is most evident when we look at results when testing on CIFAR-10. All methods drastically reduce their performance, with Binning even producing negative correlations in some cases.

Thus, we are left to conclude that image features by themselves are not the main determining factor in learning consistency of samples. This is expected as the task itself should also determine the difficulty of learning a sample: it has been seen that natural images with random labels are learned more slowly\cite{pmlr-v70-arpit17a}. What is not expected is how little they seem to generalize to out of distribution samples, even when producing simplified versions of the problem such as when using Binning and BPR. 
Future work should focus on finding other factors that explain C-Score. In particular, we believe that considering how similar are the labels for nearby samples can help explain difficulty. \cite{DBLP:conf/icml/JiangZTM21} explore this and find that such metrics only work well after training. They do not attempt to do this with ImageNet pretrained models, which we think can give valuable information.
\begin{table}[t]
\caption{Spearman Rank Correlation (SRC) for different models trained on CIFAR-100 on test sets of CIFAR-10 (out of distribution) and CIFAR-100 (in distribution). Average over 10 different seeds.}
\label{sample-table}
\vskip 0.15in
\begin{center}
\begin{small}
\begin{sc}
\begin{tabular}{llrr}
\toprule
  Dataset &      Method &  SRC &       Std. Dev \\
\midrule
 \textbf{CIFAR-100} &         \textbf{BPR} &     \textbf{0.443082} &  \textbf{0.006912} \\
 CIFAR-100 &  REGRESSION &     0.368591 &  0.011231 \\
 CIFAR-100 &      BINS-5 &     0.300274 &  0.009779 \\
 CIFAR-100 &     BINS-10 &     0.326956 &  0.008409 \\
 CIFAR-100 &     BINS-20 &     0.337483 &  0.009155 \\
 CIFAR-100 &     BINS-40 &     0.324599 &  0.008783 \\
 \midrule
  \textbf{CIFAR-10} &         \textbf{BPR} &     \textbf{0.227766} &  \textbf{0.022339} \\
  CIFAR-10 &  REGRESSION &     0.202101 &  0.014833 \\
  CIFAR-10 &      BINS-5 &     0.102341 &  0.009429 \\
  CIFAR-10 &     BINS-10 &     0.105720 &  0.006573 \\
  CIFAR-10 &     BINS-20 &     0.132004 &  0.011930 \\
  CIFAR-10 &     BINS-40 &     0.137987 &  0.010076 \\
\bottomrule
\end{tabular}
\end{sc}
\end{small}
\end{center}
\vskip -0.1in
\vspace{-3mm}
\end{table}

\section{Related Work}

Methods for ranking the difficulty of examples are standard in the Curriculum Learning literature. Traditional Curriculum Learning\cite{10.1145/1553374.1553380} uses a fixed ordering that remains unchanged during training. Self Paced Learning\cite{Kumar2010} methods derive difficulty relative to the current hypothesis of the model, usually based on a function of the model's loss. For the image domain, work has been done to define difficulty in terms of the time needed for humans to search for objects in images\cite{Ionescu2016}. Recent methods have focused on finding difficulty metrics that are based on what is empirically difficult for deep learning models. Metrics such as how often examples are forgotten\cite{DBLP:conf/iclr/TonevaSCTBG19}, learning consistency\cite{DBLP:conf/icml/JiangZTM21} and prediction depth \cite{Baldock2021} have been proposed to enable Curriculum Learning for arbitrary datasets. \cite{DBLP:conf/icml/JiangZTM21} also study distance-based and learning-speed based proxies to their C-Score method, however, both still require training on the dataset to achieve competitive results, unlike the current work which seeks to find methods to predict C-Scores on unseen datasets.
\vspace{-0.3cm}
\section{Conclusions}
\vspace{-0.2cm}
Three methods for predicting C-Score were tested: Regression, Binning and BPR. BPR proved the more successful method with the greatest Spearman Rank Correlation. All methods, however, were poor at predicting C-Scores on the same dataset as well as on a new dataset. This suggests that \textit{C-Scores are not predictable from image features alone}. Our hypothesis is that C-Scores also depend on the task, in particular, in the relation between a sample and its neighbours labels. We will attempt to verify this in future work. We also plan on including results using Inception\cite{inception} and models trained on ImageNet\cite{deng2009imagenet}, as features learned from CIFAR-100 may be too simple to transfer learn to a different dataset.
\section{Acknowledgements}
We would like to thank Carlos Aspillaga for early inspiration for this paper. This research has been financially supported by the National Center for Artificial Intelligence CENIA FB210017, Basal ANID.


\bibliography{difficulty_transfer}
\bibliographystyle{icml2022}

\newpage
\appendix
\onecolumn
\section{Contribution of LatinX Individuals}

All authors identify as LatinX. Alain Raymond-Sáez came up with the idea, developed code, experiments, analysis and wrote the paper. Julio Hurtado contributed with code, experiment suggestions and discussion. Álvaro Soto oversaw the whole research, with suggestions and discussion.

\end{document}